\documentclass{article}

\usepackage[preprint,nonatbib]{nips_2018}

\usepackage{hyperref}
\usepackage{graphicx}
\usepackage{longtable}
\usepackage{booktabs}
\usepackage{amsmath}
\usepackage{capt-of}

\title{Population Anomaly Detection \\through Deep Gaussianization}
\author{David Tolpin, dvd@offtopia.net}

\begin{document}
\maketitle

\begin{abstract}

    We introduce an algorithmic method for population anomaly
    detection based on gaussianization through an adversarial
    autoencoder. This method is applicable to detection of
    `soft' anomalies in arbitrarily distributed
    highly-dimensional data.

    A soft, or population, anomaly is characterized by a shift in
    the distribution of the data set, where certain elements
    appear with higher probability than anticipated. Such
    anomalies must be detected by considering a sufficiently
    large sample set rather than a single sample.

    Applications include, but not limited to, payment fraud trends,
    data exfiltration, disease clusters and epidemics, and
    social unrests.  We evaluate the method on several domains
    and obtain both quantitative results and qualitative
    insights.

\end{abstract}

\section{Introduction}

Divergences between anticipated and actual distribution of the
data, colloquially called data anomalies, are often analysed on
the level of individual elements of the data set: a yellow ball
in the basket where only red balls would be considered an
anomaly. A less extreme example would be a basket of red balls
with yellow spots, but with still `enough' red exposed. Here, a
ball with, say, more than 90\% of the surface covered with
yellow spots would be considered an anomaly.

However, there are cases when the anomaly can be identified by
only considering the whole data set. For example, a basket
contains red and yellow balls.  We expect the number of red and
yellow balls to be about the same. An anomaly then is five times
as many yellow balls as red balls in the basket.

\begin{figure}[h!]
    \begin{minipage}[c]{0.5\textwidth}
    \centering
    \includegraphics[width=0.7\textwidth]{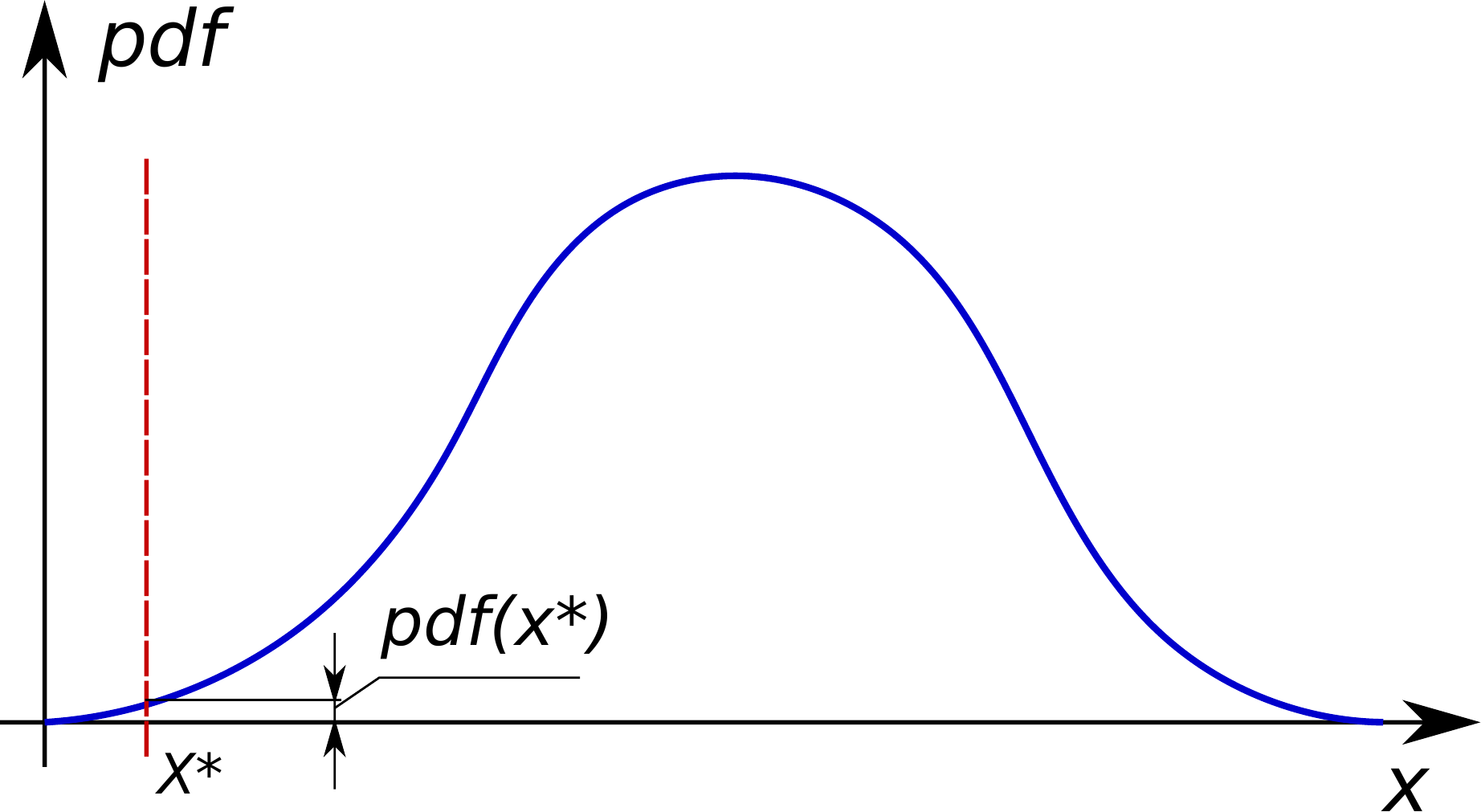}

        (a)
    \end{minipage}
    \begin{minipage}[c]{0.5\textwidth}
    \centering
    \includegraphics[width=0.7\textwidth]{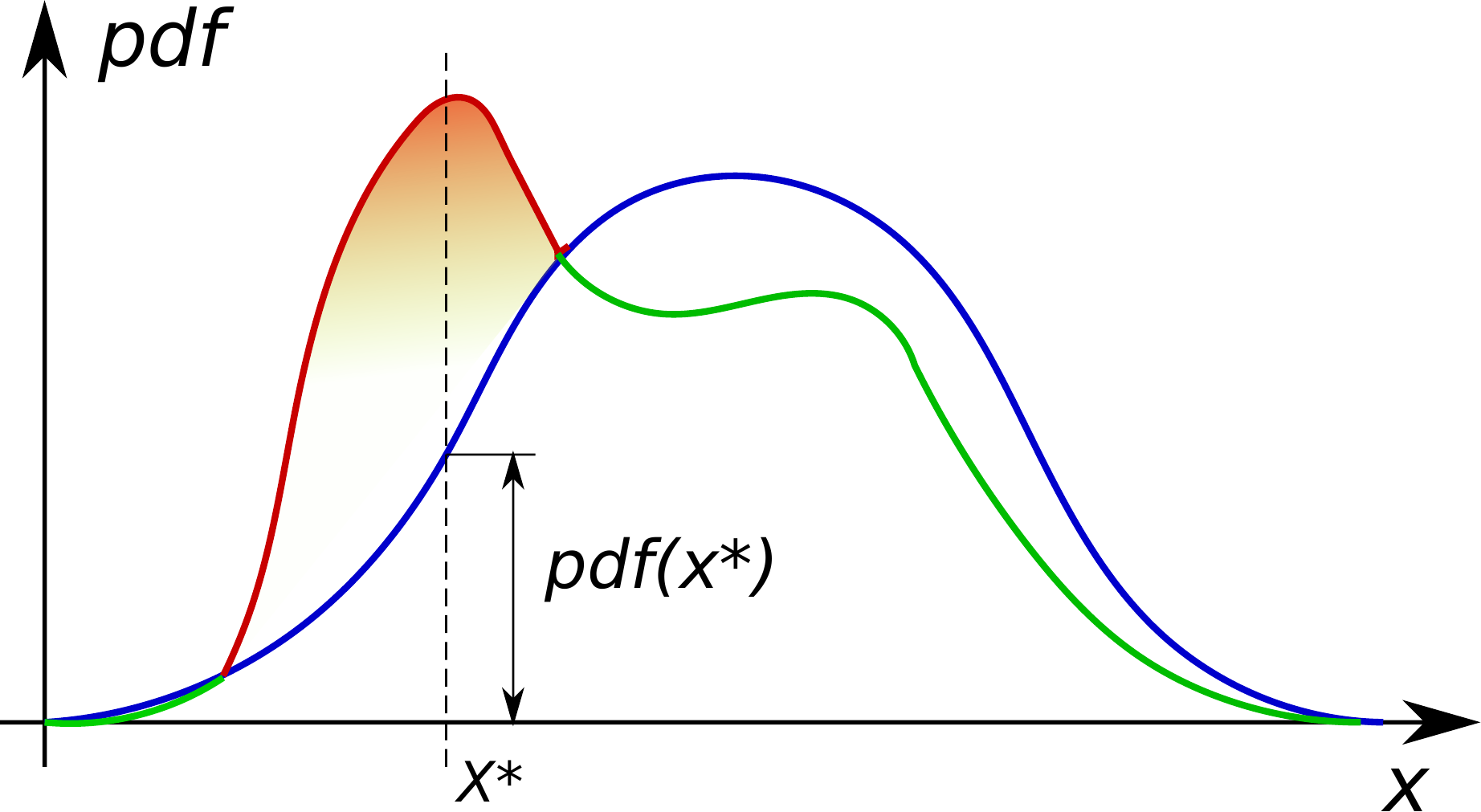}

        (b)
    \end{minipage}
    \caption{Individual (a) vs. population (b) anomaly.
    Anomalous data points may have high density w.r.t. the anticipated distribution.}
    \label{fig:popan}
\end{figure}

This work is concerned with the latter case. We call
\emph{population anomaly} a phenomenon where the distribution of
elements, rather than an individual element taken an isolation,
is abnormal.  In population anomalies, each anomalous data point
may have high probability mass or density w.r.t. the anticipated
distribution (Figure~\ref{fig:popan}).  Fraud trends in
electronic payment systems, disease clusters in public health
care, data exfiltration through network protocols are just some
examples of population anomalies.  While considering a
population anomaly, we want to evaluate the hypothesis that the
distribution underlying the data set diverges from the
anticipated distribution, and, assuming that the data set is a
mixture, obtain the probability for each sample to come either
from the regular or from the anomalous component of the mixture.

In a single dimension, the problem can be solved rather
straightforwardly, for example, by performing Kolmogorov-Smirnov
test or constructing a histogram. However, as the number of
dimensions grows, in particular in presence of complicated
interdependencies and heterogeneous data types, straightforward
brute-force approaches stop working, which is known as `the
curse of dimensionality.' Building on previous work, we propose
a method for efficient detection and ranking of population
anomalies.

\section{Problem statement}

We formulate population anomaly detection as the following machine
learning problem.

We are given a data set $S = \{x \in \mathcal{R}^k\}$ where
each sample $x$ is i.i.d. from an unknown distribution
$P=P_0$ --- the \emph{training set}.  Further on, we are given
a data set $S' = \{x' \in \mathcal{R}^k\}$ where each sample
is drawn from $P'$ which is a mixture of $P_0$ and unknown
distribution $P_1$ --- the \emph{evaluation set} with unknown
mixing probability $\gamma$. We assume that, given a sample set of
sufficient size from each of $P_0, P_1$, $P_0$ and
$P_1$ can be distinguished at any given confidence level.

We want to test the hypothesis that $S$ and $S'$ were drawn from
two different distributions $P_0$ and $P'$ and to assess the
probability $\alpha(x')$ for each sample $x'\in S'$ to be drawn
from $P_1$ rather from $P_0$.

\section{Related work}

Related work belongs to three areas of machine learning
research: population anomalies, gaussianization, and adversarial
autoencoders.

Population anomaly detection and divergence estimation is
explored in \cite{PBX+11,XPS+11}. \cite{YHL16} apply population
(group) anomaly detection to social media analysis.

Guassianization as a principle for tackling the curse of
dimensionality in high-dimensional data was first introduced in
\cite{CG01} and further developed in \cite{LCM11, EJR+06} and
other publications. Iterative algorithms involving
component-wise gaussianization and ICA were initially proposed,
with various modifications and improvements in later
publications.

Adversarial autoencoders, a deep learning architecture for
variational inference, facilitate learnable invertible
gaussianization of high-dimensional large data sets. The
architecture was introduced in \cite{MSJ+16}. The use of
autoencoders in general and adversarial autoencoders in
particular for detection of (individual) anomalies is explored
in \cite{HHW+02,SSW+17,ZP17,CSA+17}.

This work differs from earlier research in that it introduces a
black-box machine learning approach to detection and ranking of
population anomalies. The approach is robust to dimensionality
and distribution properties of the data and scalable to large
data sets.

\section{The Method}
\label{sec:method}

To handle population anomalies, we employ an
adversarial autoencoder~\cite{MSJ+16} to project the anticipated
distribution of the data set into a multivariate unit normal
distribution, that is to apply \textit{multivariate
gaussianization}~\cite{CG01} to the data. 
We then use the gaussianized representation to detect and
analyse population anomalies.

\subsection{Gaussianization}

\begin{figure}
    \centering
    \includegraphics{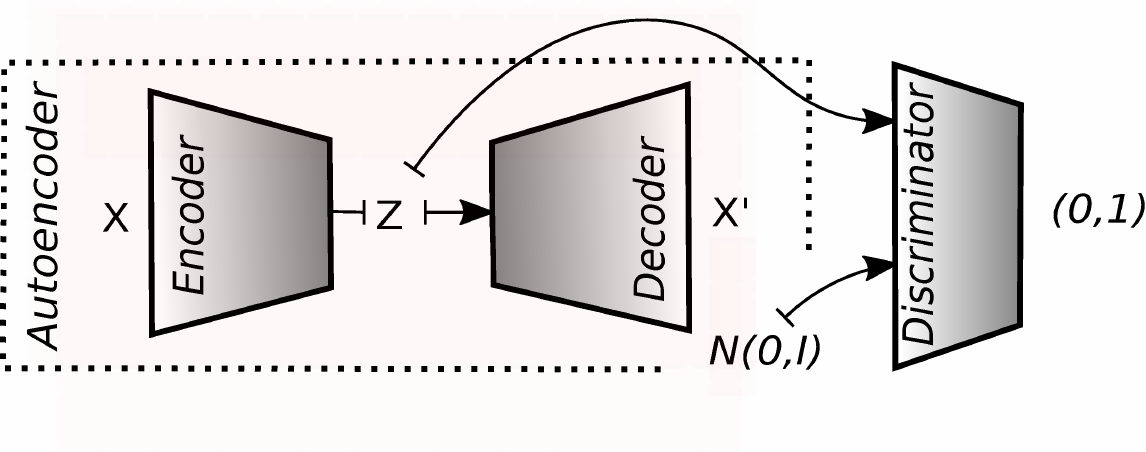}
    \caption{Adversarial autoencoder}
    \label{fig:aae}
\end{figure}

An adversarial autoencoder (Figure~\ref{fig:aae}) consists of two networks, the \emph{autoencoder}  and the \emph{discriminator}.

The autoencoder has two subnetworks, the \emph{encoder} and the
\emph{decoder}. The encoder projects the data into internal
representation, which is in our case a multivariate unit normal,
$\mathcal{N}(0, I)$. The decoder reconstructs the data sample from
a point in the space of the internal representation. The
discriminator ensures that the internal representation is indeed
normally distributed.

After training, the encoder and decoder implement \emph{invertible
gaussianization.} For every sample in the data set, a corresponding
sample from $\mathcal{N}(0, I)$ is computed by the encoder, and the
sample can be recovered by the decoder.

We train the autoencoder on the training data set, which
represents the anticipated distribution.  Then, we use the
encoder to project the evaluation data set on the space of the
internal representation. If the distribution of the evaluation
set diverges from that of the training set, the projection of
the evaluation set will diverge from unit multivariate normal
distribution.

\subsection{Detection}
\label{sec:detection}

Unit multivariate normal distribution eliminates the curse of
dimensionality because the dimensions are mutually independent.
Instead of testing the projection of the evaluation set against
the multivariate distribution, we can test the distribution
along each dimension independently, and then combine statistics
over all dimensions to detect an anomaly.

Any goodness-of-fit test assessing normality of a sample can be
used. Statistics which scale well to large sample sizes and are
sensitve to local discrepancies in the distributions should be
preferred. In our realization of the method we use the
Kolmogorov-Smirnov statistic, which allows natural probabilistic
interpretation and works sufficiently well in practice (see
Section~\ref{sec:empirical}).

For combining the statistics over all dimensions, we use a
p-norm. In simple computational evaluations $L^1$ norm worked
well enough.  When we are concerned with anomalies caused by
small intrusions or perturbations in particular, $L^\infty$,
that is, taking the maximum of statistics over axes, is a
reasonable choice. $L^\infty$ also serves as a lower bound on
the hypothesis test that $S$ and $S'$ come from the same
distribution. If the hypothesis can be rejected (that is, there
is a population anomaly in $S'$) based on a single dimension of
the gaussianized representation, then by all means the
hypothesis could have been rejected if all dimensions were
considered.

\subsection{Ranking}

In addition to testing for presence of an anomaly in the
evaluation set, we would like to rank each element of the
evaluation set by the probability to belong to the anomaly.

Here again we leverage the adversarial autoencoder. The
\emph{discriminator} component is trained to distinguish between
the projection and the unit multivariate normal distribution. We
will reuse the discriminator component to predict the anomality
of each element.

As trained during the training phase of the adversarial autoencoder,
discriminator is not yet useful for ranking the evaluation set. However,
we can take the pre-trained discriminator and \emph{train on the
evaluation set} to distinguish between the projection of the evaluation
set and samples from the unit multivariate normal distribution. The more the
evaluation set diverges from the training set, the higher will be
classification accuracy. Elements which are more likely to come from the
anomalous component ($P_1$ in the problem statement) will be
classified as such with higher confidence.

Indeed, we rank the elements of the evaluation set using the
discriminator:

\begin{enumerate}
\def\labelenumi{\arabic{enumi}.}
\item
  We project the evaluation set into the internal representation using
  the encoder trained on the training set.
\item
  We train the discriminator to distinguish between the projection of
  the evaluation set and the unit multivariate normal
  distribution, assigning label $1$ to the evaluation set and
  label $0$ to random samples.
\item
  After training, we classify the projection of the evaluation set by
  the discriminator and use the predicted label ($1$ is definitely an
  anomaly, $0$ is definitely a random sample) as the rank of anomality,
  and then backpropagate the labels to the original data.
\end{enumerate}

The discriminator is trained with binary cross-entropy loss. An
optimally trained discriminator will rank each projection $z'$
of sample $x'$ with the probability $\beta(x')$ of the
projection (and hence of the data sample) to come from $P'$. 
$\beta(x')$ can be used to estimate the probability $\alpha(x')$
of $x'$ to come from $P_1$. Indeed, denoting the densities
of $P_0$ and $P_1$ as $f_0$ and $f_1$ correspondingly, and the
ratio $\frac {f_1(x')} {f_0(x')}$ as $\varphi(x)$ we obtain:

\begin{minipage}[c]{0.6\textwidth}
\begin{align}
    \nonumber
    \alpha(x') & =  \frac {\gamma \varphi(x')} {1 - \gamma + \gamma \varphi(x')},\quad \beta(x') = \frac {1 - \gamma + \gamma \varphi(x')} {2 - \gamma + \gamma \varphi(x')} \\ \nonumber
    \\
    \alpha(x') &\approx  2 - \frac 1 {\beta(x')} \quad \mbox{ for } \gamma \ll 1,\,\beta(x') \ge \frac 1 2 
\end{align}
\end{minipage}
\begin{minipage}[c]{0.4\textwidth}
    \centering
    \includegraphics[width=0.8\textwidth]{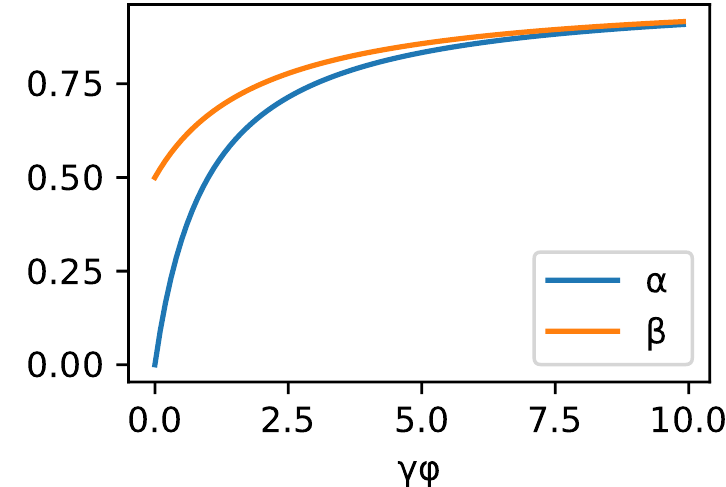}
    \captionof{figure}{$\alpha$ and $\beta$ vs. $\gamma\varphi$
    for $\gamma \ll 1$.}
    \label{fig:alpha-beta}
\end{minipage}

$\alpha(x')$ increases with $\beta(x')$ as function of $\gamma
\varphi(x')$. For $\gamma \ll 1$, which is the case in anomalies,
$\alpha(x')$ and $\beta(x')$ as functions of $\gamma \varphi(x')$
are shown in Figure~\ref{fig:alpha-beta}. $\alpha = \frac 1 2$,
that is, the density of anomalous samples being equal to the
density of regular samples, corresponds to $\beta \approx \frac 2 3$.

\subsection{Method Outline}
\label{sec:outline}

Let us now summarize the algorithmic steps constituting the method:

\begin{itemize}
    \item Training:
        \begin{enumerate} 
            \item Train the adversarial autoencoder on the training set.
        \end{enumerate}
    \item Detection:
        \begin{enumerate}
            \item Project the evaluation set on the internal representation space
                using the \textit{encoder}.
            \item Compute KS statistics for each dimension of the projection.
            \item Combine the compute statistics over all dimensions using a p-norm
                (e.g. take the maximum KS statistic) and use the combined value to
                test whether an anomaly is present.
        \end{enumerate}
    \item Ranking:
        \begin{enumerate}
            \item Train the \textit{discriminator} to distinguish
                between the projection of the evaluation set and
                random samples from the unit multivariate normal
                distribution.
            \item Classify the evaluation set using the trained
                discriminator network.
            \item Sort the elements in the evaluation set
                according to the rank assigned by the
                classifier.
            \item Report elements with the highest ranks as the
                most `surprising' ones, i.e. those most likely
                to belong to an anomaly.
        \end{enumerate}
\end{itemize}

\section{Empirical Evaluation}
\label{sec:empirical}

In the empirical evaluation that follows we evaluate the method
on three domains of different structure and from different
application areas. In all cases, point-based anomaly detection
cannot be applied to detect the anomalies as the probability of
each individual element belonging to the anomaly is as high as or
higher than of some of the regular elements in the training set.

However, unusually high probabilities of the anomalous elements in the
evaluation set indicate the anomaly which is detected by the
introduced method for population anomaly detection.

\subsection{Credit Card Payments}\label{credit-card-payments}

We were provided with a data set of credit card transaction data
over a month. The data set contains $\approx 2$ million
transactions. We divided the data into 168 buckets, for each hour of each
day of the week.  For each bucket, a separate model is trained. A
data record consists of 14 fields of both continuous
(transaction amount, conversion rate) and categorical (country,
currency, market segment, etc.) types. In the expanded form, each record
is represented by a 415-element vector.  8-dimensional internal
representation was used. Since the data is a mixture of
continuous and categorical values, mean squared error loss was
used as the reconstruction loss of the autoencoder.

We use the method to compare different hours of day and
different days of week. There are $(24 \cdot 7)^2 = 28224$
possible combinations of model and data, to which we apply
the method.

\subsubsection{Detection}

\begin{figure}
\begin{minipage}[c]{0.6\textwidth}
    \centering
\includegraphics[width=0.7\textwidth]{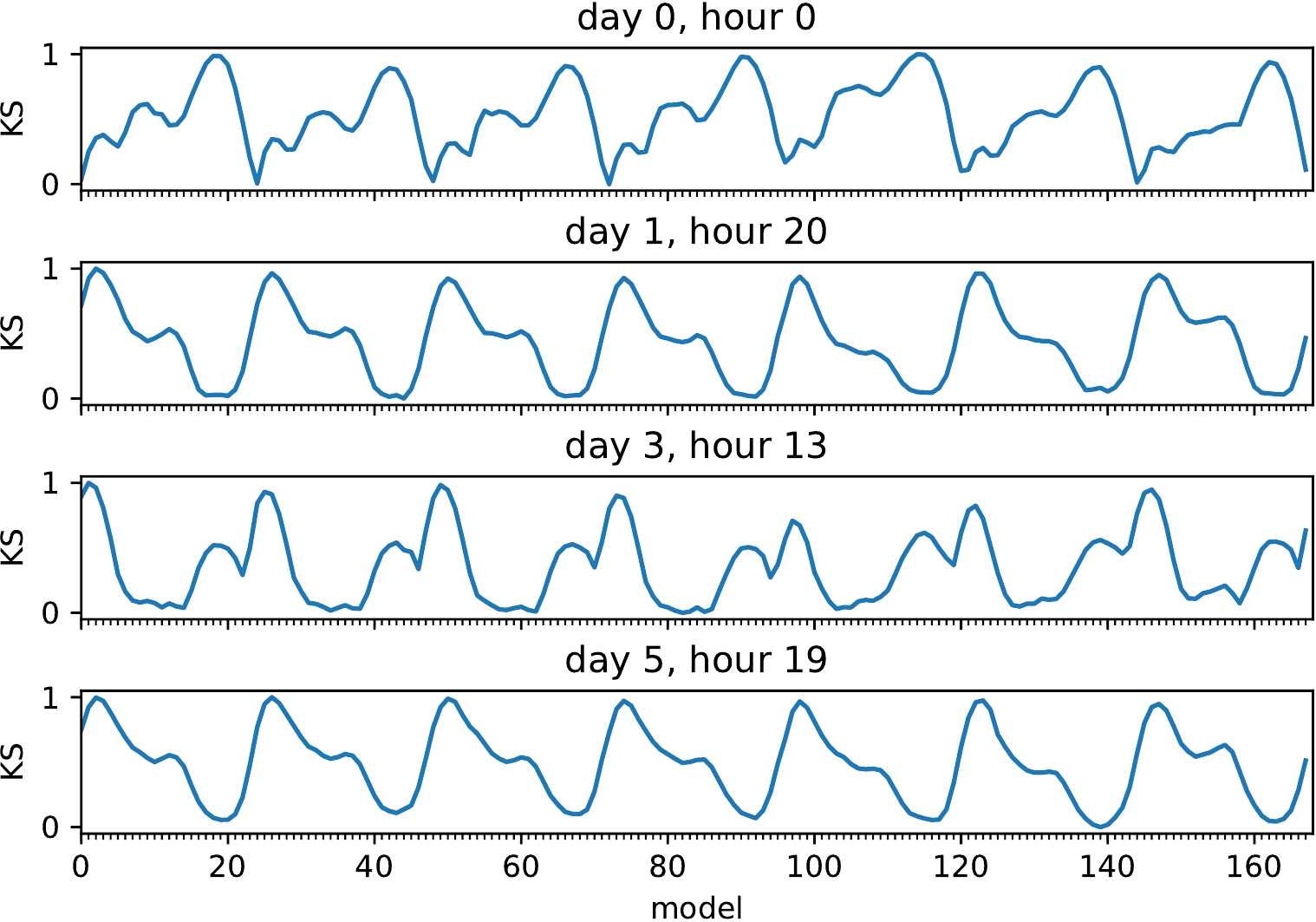}

a.
\end{minipage}
\begin{minipage}[c]{0.4\textwidth}
    \centering
\includegraphics[width=0.75\textwidth]{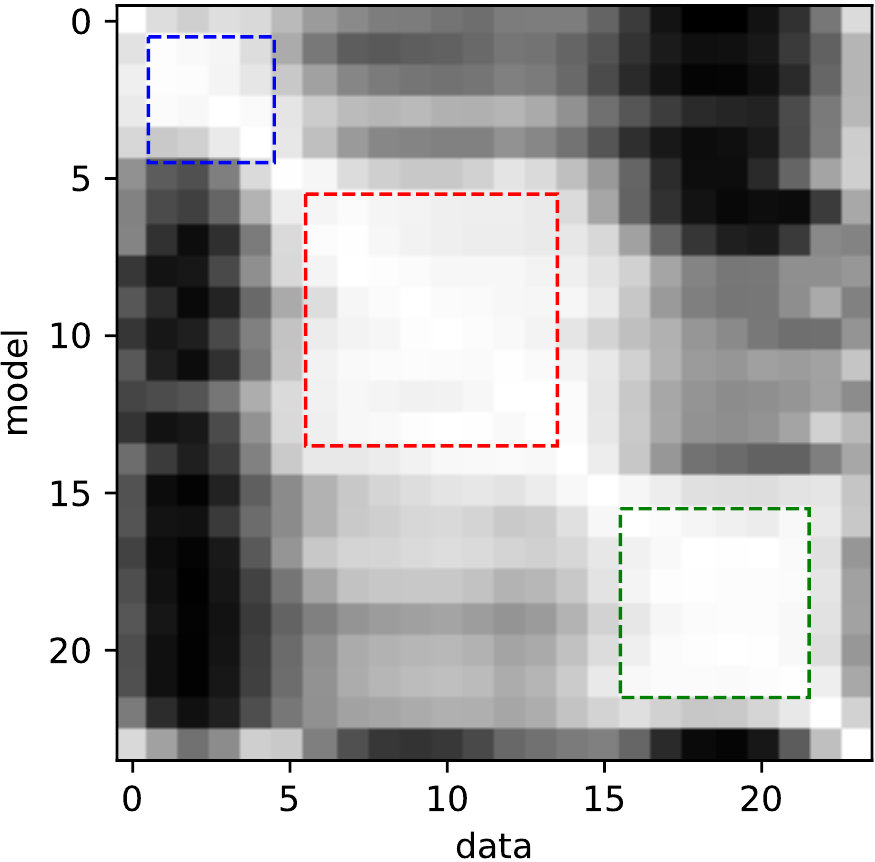}

b.
\end{minipage}
\caption{Novelty in credit card transaction data.}
\label{fig:cc}
\end{figure}

Figure~\ref{fig:cc} presents some of results of quantifying 
anomality (novelty) between different hours of the week.
The hours are per the Pacific Time Zone.

Figure~\ref{fig:cc}a shows novelty of each hour over a week
relative to a particular hour's model, for 4 randomly selected
hours. 
\begin{itemize}
    \item Same hours on different
        days, even as different as Monday and Sunday, have similar
        distributions. 
    \item Weekdays are more similar to each other 
        than a weekday and a weekend.
\end{itemize}

Figure~\ref{fig:cc}b shows relative novelty in each pair
of hours over a single day (Wednesday). The lighter the square,
the more similar the model and the data hours are.  There are
three regions of similarly looking hours (appearing as light
squares on the plots). We marked these regions by dashed colored
boxes. Ranking of transactions in hours belonging to different
regions (see Section~\ref{sec:cc-ranking}) suggests
geographical interpretation of peak activities:
\begin{itemize}
\item
    1 -- 4 (blue box) --- Europe and Middle East
\item
    6 -- 14 (red box) --- Americas
\item
    16 -- 23 (green box) --- Asia-Pacific
\end{itemize}

\subsubsection{Ranking}
\label{sec:cc-ranking}

We consider top-ranked transactions from several combinations of
data and model hours. For simplicity, only the sender and the
receiver country are shown here, however other fields may have
also affected the ranking.

\paragraph{Sanity check --- same hour}

First, we make sure that the transactions are not surprising when they
come from the model's bucket (0.5 is the neutral rank):

\begin{minipage}{\textwidth}
    \centering
\textbf{03:00 on Monday}

\begin{minipage}{0.45\textwidth}
    \centering
    \textbf{Most surprising}

\begin{tabular}{c c l l}
\textbf{rank} & \textbf{sender} & \textbf{receiver} & ... \\ \hline
0.551 & IT & DE & ... \\
0.550 & DE & DE & ... \\
0.549 & AT & DE & ... \\
0.548 & DE & DE & ...  \\
0.547 & DE & DE & ... \\
\end{tabular}
\end{minipage}
\begin{minipage}{0.45\textwidth}
    \centering
    \textbf{Least surprising}

\begin{tabular}{c c l l}
\textbf{rank} & \textbf{sender} & \textbf{receiver} & ... \\ \hline
0.416 & DK & US & ... \\
0.414 & US & US & ... \\
0.408 & US & US & ... \\
0.406 & US & US & ... \\
0.386 & HK & IE & ...
\end{tabular}
\end{minipage}
\end{minipage}

The highest probability is $\approx 0.55$ and the lowest is
$\approx 0.39$ which is a rather narrow range of surprise, as
expected.

\paragraph{Different hours within the same day}

Comparing different hours on the same day helps give
interpretation to different similarity regions (colored boxes)
in Figure~\ref{fig:cc}b.

\begin{minipage}{0.5\textwidth}
    \centering
\textbf{12:00 vs. 3:00 on Wednesday}

\begin{tabular}{c c l l}
\textbf{rank} & \textbf{sender} & \textbf{receiver} & ... \\ \hline
0.999 & US & US & ... \\
0.930 & US & US & ... \\
0.900 & US & US & ... \\
0.886 & US & US & ... \\
0.884 & US & US & ...
\end{tabular}
\end{minipage}
\begin{minipage}{0.5\textwidth}
    \centering
\textbf{3:00 vs. 12:00 on Wednesday}

\begin{tabular}{c c l l}
\textbf{rank} & \textbf{sender} & \textbf{receiver} & ... \\ \hline
0.724 & GB & GB & ... \\
0.718 & IT & IT & ... \\
0.718 & IT & IT & ... \\
0.717 & GB & GB & ... \\
0.716 & IT & IT & ...
\end{tabular}
\end{minipage}

The most surprising transactions at 12:00 on Wednesday compared
to 3:00 are payments within the US. When ranked in the opposite
direction (ranking is \textbf{not symmetric}), the most
surprising transactions at 3:00 compared to 12:00 are
payments within Europe. Let's now check the evening hours:

\begin{minipage}{\textwidth}
    \centering
\textbf{22:00 vs. 12:00 on Wednesday}

\begin{tabular}{c c l l}
\textbf{rank} & \textbf{sender} & \textbf{receiver} & ... \\ \hline
0.706 & HK & TW & ... \\
0.705 & AU & AU & ... \\
0.703 & AU & AU & ... \\
0.703 & AU & AU & ... \\
0.702 & HK & HK & ...
\end{tabular}
\end{minipage}

At 22:00 the most surprising transactions relative to 12:00 are
those within the Far East.

\paragraph{Same hour, different days}

Same hours on different days are generally similar, but we saw that
weekends are different from weekdays. Let's try to explain some of the
differences:

\begin{minipage}{\textwidth}
    \centering
\textbf{2:00 on Sunday vs. on Monday}

\begin{tabular}{c c l l}
\textbf{rank} & \textbf{sender} & \textbf{receiver} & ... \\ \hline
0.705 & DE & DE & ... \\
0.704 & DE & DE & ... \\
0.699 & DE & DE & ... \\
0.699 & DE & DE & ... \\
0.699 & DE & DE & ...
\end{tabular}
\end{minipage}

At 2:00 on Sunday the most surprising transactions relative to
Monday 2:00am are certain payments within Germany (probably involving
other attributes).

\subsection{London Crime Data}

The Kaggle data set of London Crime~\cite{KaggleLondonCrime}
contains $\approx 6.5$ million of unique crime cases for
years 2008--2016. Each crime case record contains the crime
category, the borough were the crime happened, and the year
and month of the event. We divided the data into 9 buckets,
a bucket per year. In the expanded form, each record is
represented by a 78-dimensional vector. 8-dimensional internal
representation was used. All fields are categorical, hence 
binary cross-entropy loss was used as the reconstruction loss
of the autoencoder.

\subsubsection{Detection}

\begin{figure}
\begin{minipage}[c]{0.4\textwidth}
\centering
\includegraphics[width=0.75\textwidth]{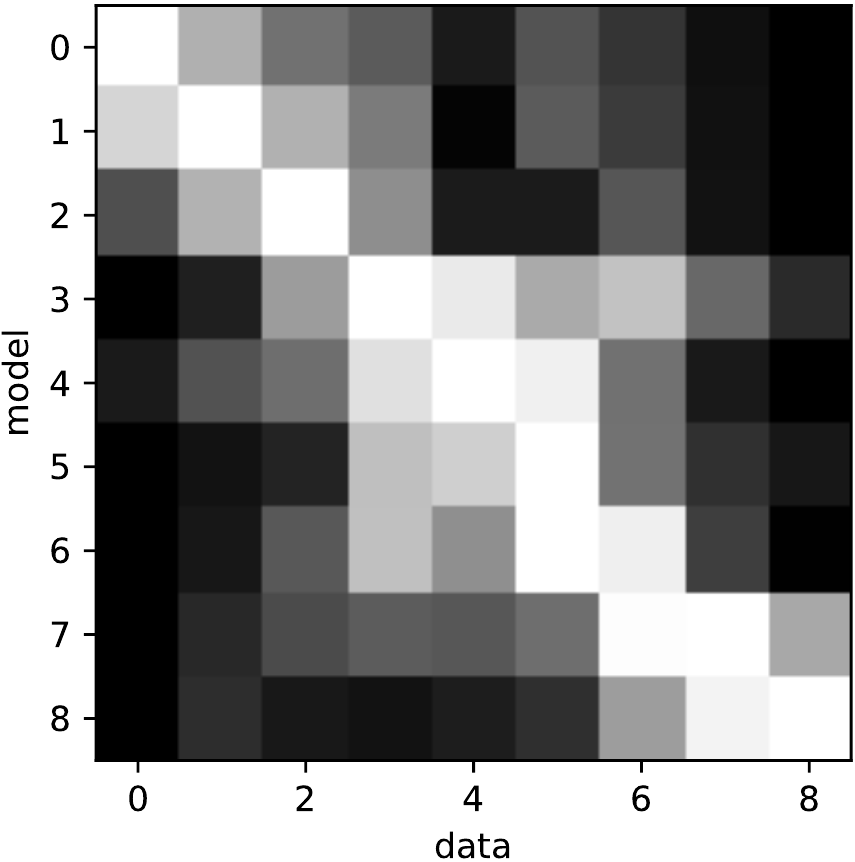}

a.
\end{minipage}
\begin{minipage}[c]{0.6\textwidth}
    \centering
\includegraphics[width=0.7\textwidth]{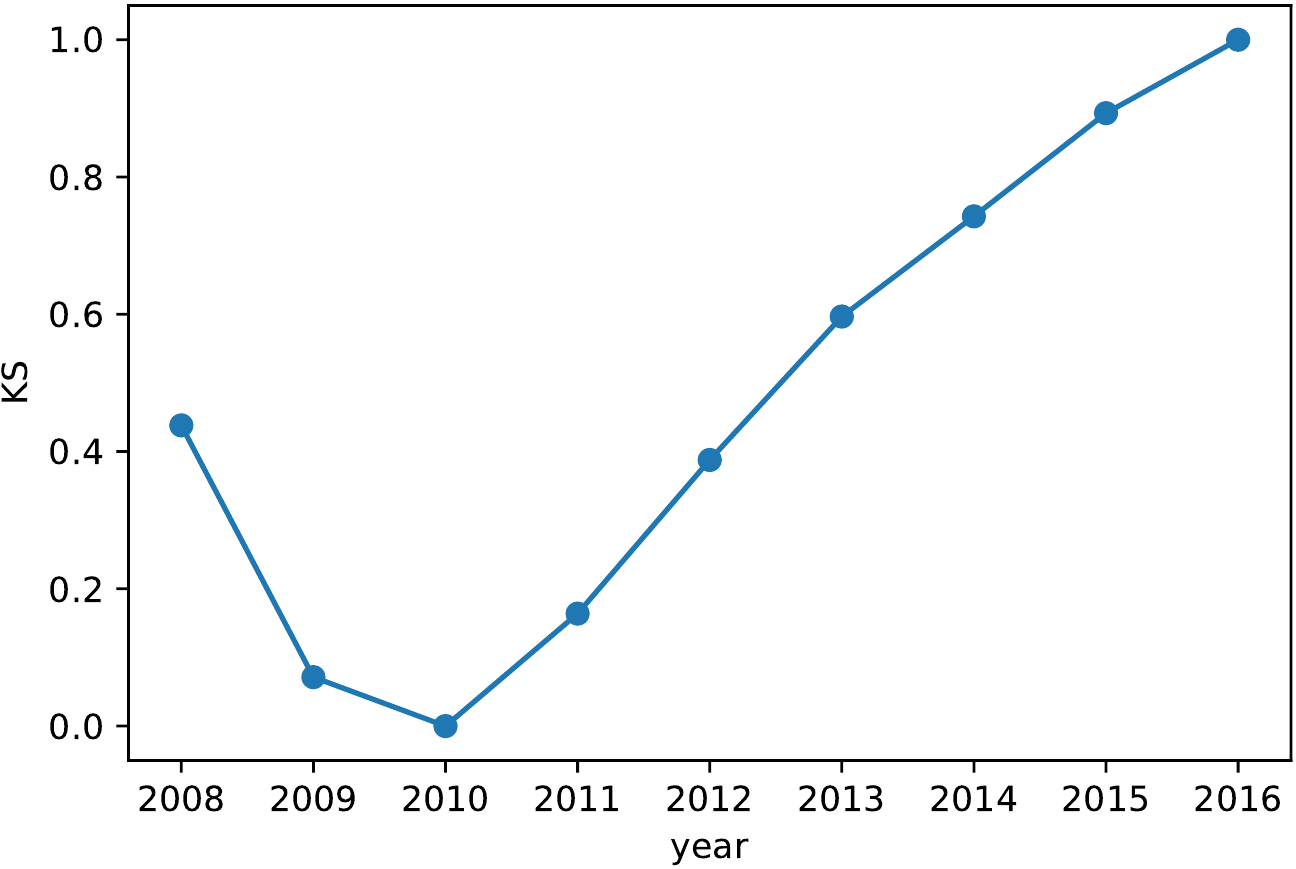}

b.
\end{minipage}
    \caption{Novelty in London crime data.}
    \label{fig:crime}
\end{figure}

Figure~\ref{fig:crime} presents results of quantifying novelty
between different years.

Figure~\ref{fig:crime}a shows relative novelty for each pair of years.
The lighter the square the more similar the years are.
Subsequent years are similar to each other, the further apart
the years the greater is the mutual novelty.

Figure~\ref{fig:crime}b shows novelty (maximum KS statistic over
the dimensions of the internal representation) of each year
relative to the model trained on data of all years. Years
2010-2011 appear to be the closest to the overall distribution
of crimes, with years at the beginning and the end of the year
range farther apart.  Year 2016 is much further from the overall
distribution than year 2008 though.

\subsubsection{Ranking}

To illustrate insights which can be obtained through ranking of
anomalous records we compare the first and the last year in the
span to each other, as well as the overall distribution to the
last year.

In year 2008 compared to 2016 the highest ranked records
compared to year 2016 are theft from motor vehicle in several
boroughs. This can be interpreted as that the frequency of this
crime decreased in London by 2016.

\begin{minipage}{\textwidth}
    \centering
\textbf{2008 vs. 2016}

\begin{tabular}{c c l l}
\textbf{rank} & \textbf{month} & \textbf{category} & \textbf{borough} \\ \hline
0.807 & 10 & Other Theft & Westminster \\
0.743 & 2 & Theft From Motor Vehicle & Islington \\
0.721 & 2 & Theft From Motor Vehicle & Wandsworth \\
0.720 & 2 & Other Theft & Haringey \\
0.717 & 2 & Theft From Motor Vehicle & Hammersmith and Fulham
\end{tabular}
\end{minipage}

In 2016 compared to 2008 the highest ranked record is of
harassment. Note that harassment is not an outlier in 2008 ---
5\% of reported crimes are harassment, compared to 11\% in 2016.
Still, harassment records appear to constitute the greatest
novelty in 2016.

\begin{minipage}{\textwidth}
    \centering
\textbf{2016 vs. 2008}

\begin{tabular}{c c l l}
\textbf{rank} & \textbf{month} & \textbf{category} & \textbf{borough} \\ \hline
0.732 & 7 & Harassment & Newham \\
0.725 & 1 & Harassment & Lambeth \\
0.722 & 2 & Harassment & Hillingdon \\
0.722 & 5 & Harassment & Hounslow \\
0.719 & 1 & Harassment & Harrow
\end{tabular}
\end{minipage}

Comparing all year's data to the model of 2016 we find that the
highest ranked records are of assault with injury in central
boroughs of London. That can be interpreted as that that
particular crime was frequent in central London, but the
frequency decreased by 2016.

\begin{minipage}{\textwidth}
    \centering
\textbf{all years vs. 2016}

\begin{tabular}{c c l l}
\textbf{rank} & \textbf{month} & \textbf{category} & \textbf{borough} \\ \hline
0.800 & 10 & Other Theft & Westminster \\
0.744 & 7 & Assault with Injury & Westminster \\
0.720 & 7 & Assault with Injury & Kensington and Chelsea \\
0.716 & 7 & Assault with Injury & Lambeth \\
0.713 & 5 & Personal Property & Westminster
\end{tabular}
\end{minipage}

\subsection{DNS-based data exfiltration}

We applied the method to detection of DNS-based data
exfiltration. The CAIDA UCSD DNS Names Dataset~\cite{CAIDA} was
used.

We emulate data exfiltration by replacing the last component of
domain in a certain fraction of the data set with a sequence of
characters sampled from characters permitted in domain names
(uppercase and lowercase letters, as well as digits and the
dash). For example, \texttt{foobar.example.com} might be
replaced with \texttt{AsdR5t.example.com}. This method
approximates the distribution of encoded data, while still
keeping the distribution of domain \emph{lengths} unaffected.
0.1\%, 1\%, and 10\% of the entries in the evaluation data set
are replaced with entries emulating data exfiltration.
Only domain names are considered for machine learning.  The
domain names were mapped to 64-dimensional (by the number of
allowed characters) vectors of character counts. 4-dimensional
internal representation was used.  Mean squared error loss was
used as the reconstruction loss of the autoencoder.

\begin{figure}
\begin{minipage}[c]{0.5\textwidth}
    \centering
\includegraphics[width=0.7\textwidth]{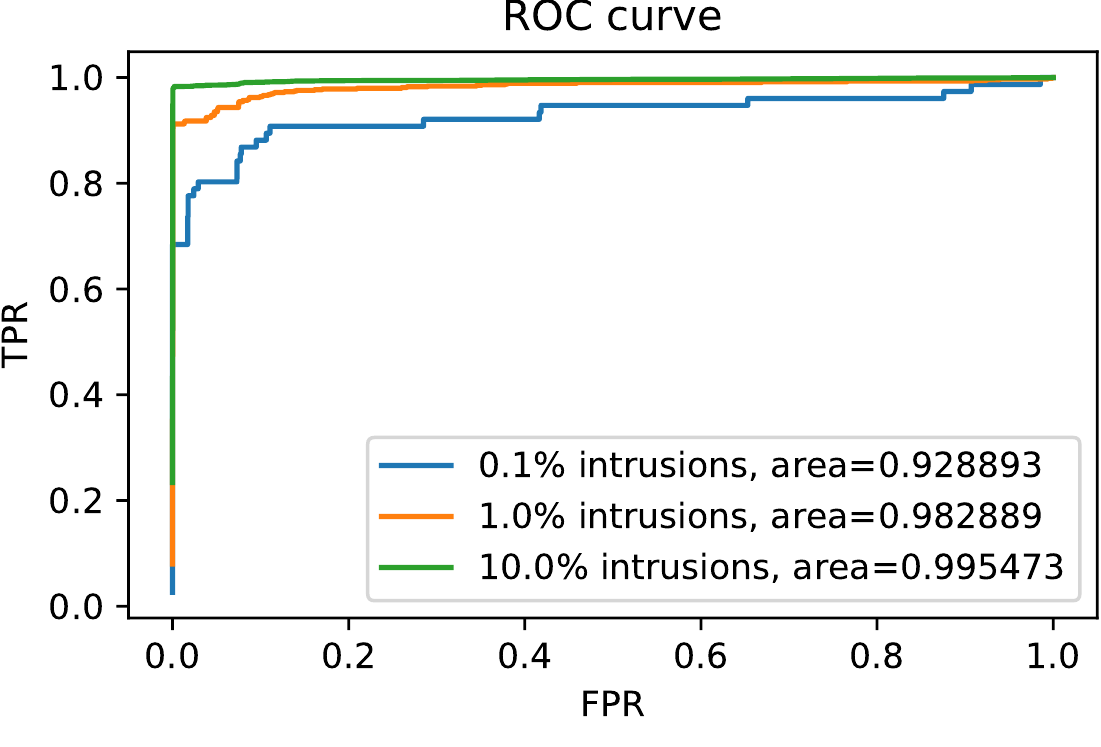}

a.
\end{minipage}
\begin{minipage}[c]{0.5\textwidth}
    \centering
\includegraphics[width=0.7\textwidth]{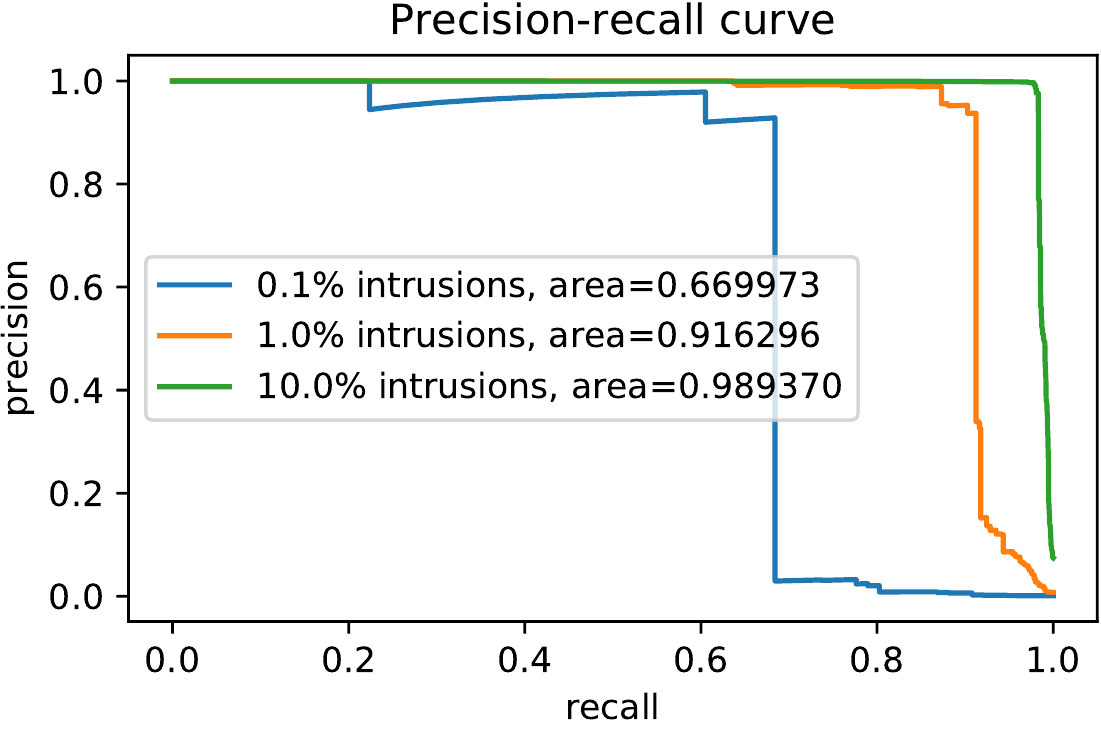}

b.
\end{minipage}
    \caption{ROC (a) and Precision-Recall (b) curves of DNS exfiltration
    detection.}
    \label{fig:dns-roc-prc}
\end{figure}

Figure~\ref{fig:dns-roc-prc} shows the ROC and precision-recall
curves histograms of exfiltration detection.

Note that the classification accuracy (for the given amount of training
budget) increases as the number of anomalous entries goes up, unlike in
methods which rank every sample individually. This self-boosting is a
useful feature of the proposed method: the more severe the attack,
the higher is the ranking accuracy.

\section{Discussion}

We described a method for detecting and quantifying population anomalies
in high-dimensional data and evaluated the method on several application
domains. An anomaly, or novelty, in the data is an unusually
high probability of occurrence of certain elements. Individual
anomalies are commonly detected based on low probability of the
elements relative to the anticipated distribution, which is
sufficient but not necessary condition of anomality.  Elements
of population anomalies may still have relatively high probability.

Population anomalies and methods of their detection have been
subject of earlier research, however the introduced method offers
a black-box approach to population anomaly detection and is
robust to data set sizes and data types and distributions. One
challenge for any population anomaly detection method introduced
so far which still needs to be addressed is \textit{explanation}
--- summary characterization of the anomaly instead of
just presenting most anomalous samples. Our method may 
be a good foundation for addressing this challenge, by allowing
augmentation and reconstruction of anomalies from the internal
representation, a subject for future research.

\bibliographystyle{plain}
\bibliography{refs}

\end{document}